\pdfoutput=1
\def\year{2017}\relax
\documentclass[letterpaper]{article} 
\usepackage{aaai17}  
\usepackage{times}  
\usepackage{helvet}  
\usepackage{courier}  
\usepackage{url}  
\usepackage{graphicx}  
\usepackage{color}
\usepackage{cancel}
\usepackage[normalem]{ulem}
\usepackage[linktocpage=true,breaklinks]{hyperref} 
\usepackage{amsmath}
\usepackage{amsfonts}
\usepackage{soul} 

\nocopyright

\begin{document}
%
\title{Modeling Epistemological Principles for Bias Mitigation in AI Systems:\\ An Illustration in Hiring Decisions}
\author{Marisa Vasconcelos, Carlos Cardonha, Bernardo Gon\c{c}alves\\
IBM Research Brasil\\
Rua Tut\'oia 1157, S\~ao Paulo -- SP 04007-900\\
}


\maketitle
\begin{abstract}
Artificial Intelligence (AI) has been used extensively in automatic decision making in a broad variety of scenarios, ranging from credit ratings for loans to recommendations of movies. Traditional design guidelines for AI models focus essentially on accuracy maximization, but recent work has shown that economically irrational and socially unacceptable scenarios of discrimination and unfairness are likely to arise unless these issues are explicitly addressed. This undesirable behavior has several possible sources, such as biased datasets used for training that may not be detected in black-box models.  After pointing out connections between such bias of AI and the problem of induction, we focus on Popper's contributions after Hume's, which offer a logical theory of preferences. An AI model can be preferred over others on purely rational grounds after one or more attempts at refutation based on accuracy and fairness. Inspired by such epistemological principles, this paper proposes a structured approach to mitigate discrimination and unfairness caused by bias in AI systems. In the proposed computational framework, models are selected and enhanced after attempts at refutation. To illustrate our discussion, we focus on hiring decision scenarios where an AI system filters in which job applicants should go to the interview phase.
\end{abstract}



\section{Introduction}

There is some common understanding today that artificial intelligence (AI) systems should be accountable and behave reasonably from an ethical point of view, e.g., see \cite{greene2016embedding}. The widespread use of black-box AI algorithms in all sorts of decision-making systems, from credit scores 
to recommendation of products,
may have unpredictable and/or potentially destructive consequences on people's lives \cite{ONeil:2016:WMD}. 
The challenges related with the way AI models are created and over-trusted have caught the attention of the research community across several areas. 

AI decision-making pipelines can be roughly represented schematically along these lines: Human bias $\to$ [Data $\to$] Algorithm $\to$ AI system $\to$ Decision making.
Undesirable  behavior of AI systems may be a consequence of human bias, which may be (unconsciously) unfair and affect significantly the output of these systems.
Namely, whether
they have embedded a hypothesis formulated by humans (say, a decision tree), or generates hypotheses automatically from training datasets, 
(human) biases may arise in the process and can be further absorbed, propagated, and amplified at scale by the resulting algorithms~\cite{barocas2014datas}.  

These challenges have been observed in several scenarios over the last few years~\cite{surveyDataBias}, frequently with considerable popular attention~\cite{propublica}. An important scenario is algorithmic hiring decisions, in which AI systems are used to filter in and out job applicants in initial screening steps. A first problem is that the historical data of a job applicant may not be all relevant or adequate to be used for the purpose of finding a good hire. Second, even new types of data are now being used (e.g., posts in a social network), whose relevance and adequacy from an ethical perspective can be questionable as well. Third, the replacement of several human decision makers, specially when they  have different and complementary views, for a single decision-making algorithm, may imply in loss of diversity for the hiring decision process. For all these reasons, building AI systems that are both economically- and ethically-acceptable for hiring decisions
is a challenging task. 

Motivated by these questions of practical interest, the research community has started to address such bias-of-AI challenges.  For example, when a structured dataset is available so that an AI decision function can learn from it, one can also design an oversight process to explicitly seek (mine) strong correlations (say, in the form of association rules) over sensitive attributes (e.g., race). This may conveniently unravel prejudice (e.g., a given race implies a good hire) that, once found, can be mitigated before the system is deployed. In a more general sense, AI rules or patterns may be learned from a classical knowledge engineering process as well (say, by interviewing a few domain experts), as opposed to machine learning from an existing dataset. In any case, there is bias in learning from examples, and it is implicitly related with the long-lived problem of induction in epistemology. There are several relevant philosophers that brought contributions to the problem --- for an overview, we refer the reader to \cite{vickers2016}. However, two of the outstanding figures are David Hume and Karl Popper, and their core contributions will suffice for the purpose of this paper. 

Consider an empirical regularity or law that has been learned from some portion of the world, say, ``All graduate white males are good hires.'' After Hume's contributions, Popper has established that no finite amount of data, however large it may be, is enough to rationally justify such a law. Nonetheless, whereas Hume ended up in skepticism about knowledge, either scientific or commonsense, Popper has devised on top of Hume's ideas a view of science centered on conjectures and refutations. This view offers a \emph{logical theory of preferences}, which enables one to select a hypothesis over another on purely rational grounds --- how they withstand attempts at refutation. Interestingly, Popper's view can be fruitfully seen as a learning theory \cite{popper1984}, which will suit well into an AI system.

In this paper we will leverage on Popper's epistemological principles and borrow contributions from the (theoretical) computer science and optimization literature towards a two-level framework for bias mitigation in AI systems. 
On its first (higher) level, it 
enables exploring a portfolio of two or more different decision functions (conjectured hypotheses in the sense of Popper). The quality of a decision function is evaluated according to  its accuracy.\footnote{We consider the basic definition of accuracy, in which system outputs are compared against their ground truth values.}
The best function is \emph{selected} with higher probability whereas reasonable alternative functions are also being constantly considered and evaluated. The second level of the framework focuses on the \emph{enhancement} of decision functions that turn out \emph{not} to be satisfactory for the target problem (after a number of attempts at refutation), either due to low accuracy or to unfairness.\footnote{Unfairness is characterized by unacceptable correlations between system outputs and sensitive attributes (e.g., race).}
For scenarios where decision functions can be reformulated (i.e., they are not black-box models), we point to connections between the  enhancement procedure and 
cut-generation procedures, originated from the fields of constrained logical programming and optimization in general.


The paper is structured is as follows. In \S\ref{sec:scenario} we briefly describe the hiring decisions scenario we will use for illustration. In \S\ref{sec:inductive_bias} we introduce connections between the bias of AI and the problem of induction. We review Hume's contributions and focus on Popper's epistemological principles of conjectures and refutations and their connections with a learning theory to be built into an AI system. In \S\ref{sec:example} we present an example in the hiring scenario in terms of a simple yet clear data model. In \S\ref{sec:model} we present the proposed computational framework in some detail, illustrated in the same scenario. In \S\ref{sec:related-work} we comment on bias-of-AI related work. Finally, in \S\ref{sec:conclusions} we provide conclusions and discuss future work.

\section{Use Scenario: Hiring Decision Algorithms}
\label{sec:scenario}


Hiring processes have a long tradition of explicit and implicit human biases, which sometimes may lead to discriminatory consequences. There are several kinds of bias that may end up influencing a hiring decision. One example is confirmation bias of first impression, i.e., when recruiters make a first assessment in the beginning of the interview and spend the rest of the time looking for reasons to support their initial impression~\cite{wired}. Mitigation actions have been introduced in the form of federal laws in order to 
 protect job applicants against discrimination based on age, race, color, religion, sex, and national origin. However, hiring discrimination is difficult to uncover, since answers to questions can be inferred from indirect questions (e.g., asking the applicant when did she graduate can be used to infer her age), leading to ethical and privacy issues.

Recently, companies started to use AI algorithms to filter job applications in their entry-level hiring process. Compared to the traditional process, such automatic decision has been claimed as a strategy to minimize human bias in the process eventually \cite{business}, increasing diversity and cost-efficiency (e.g., hiring selection process speed up).  Algorithms are helping companies to model historical patterns of hiring from data describing ``high-performance''' or ``ideal'' employees so that they can find candidates with similar profiles. However, bias coming from the traditional hiring process is still coded in the data used to train those systems. Thus, extracting insights about whether a person is a good match for a job from structured data is not trivial, and this task becomes even harder when associated with unstructured data. For instance, 
some startups are offering services that go from brain games that assess capabilities as focus, risk-taking, and memory to platforms that analyze audio (e.g., voice intonation), video (e.g., body language, blink frequency), text (e.g., keywords), and social data posts (from social networks to websites) to construct a psychological profile of the candidate and compare it  against the culture of the hiring company. It is not clear whether this kind of data is a valid proxy for the ability of individual to perform the job, and generalizations can be even more complicated if significant difference in cultural backgrounds exists.




%

Finally, by using only the perspective of a single decision maker algorithm, we may lose the individuality or the different criteria of multiple decision makers.  For instance,  different members of a hiring committee may assign different weights to relevant aspects describing a job applicant, enabling then a fair assessment of her. That may not be captured well by a single model or algorithm. 

In the next section, we show how problems that have been already discussed by philosophers of science are related to some of the current challenges faced in the development of AI systems. We believe that this is a two-way process, in the sense that the present discussion  has potential to bring material of interest to  traditional epistemological discussions.

\section{Bias of AI and the Problem of Induction}
\label{sec:inductive_bias}
A commonsense dictionary definition for `bias' is: (sense one) ``the action of supporting or opposing a particular person or thing in an unfair way, because of allowing personal opinions to influence your judgment''; and (sense two) ``the fact of preferring a particular subject or thing'' \cite{dictcambridge2}. While the former indicates an unfair judgement, the latter refers to a \emph{preference}. In fact, every decision making process is about arriving at a reasonable preference (bias, in the second sense). 



Avoiding bias in the first sense (prejudice) is a challenge from an ethics point of view. Yet, as we will see in this section in a summarized way that suffices for the purpose of this paper, it has strong connections with the long-lived problem of induction in epistemology. It has been critically examined by David Hume in the 18th century \cite{hume1777}, and then by Karl Popper last century. After \citeauthor{hume1777}'s remarks, \citeauthor{popper1953} formulated two variants he called the \emph{logical} and the \emph{psychological} problems of induction \shortcite[p.~107-8]{popper1953}:%
\footnote{Cf. also \cite[Chapter ``Conjectural Knowledge'']{popper1972}.} 

\begin{itemize}
  \item \emph{Logical}. Are we rationally justified in reasoning from repeated instances of which we have had experience to instances of which we have had no experience? Hume's answer is no, we are not justified, however great the number of repetitions may be; and this is so for \emph{certain} belief, as well as for \emph{probable} belief. Instances of which we have had experience do not allow us to reason or argue about the \emph{probability} of instances of which we have had no experience, any more than to the \emph{certainty} of such instances. 

\item \emph{Psychological}. How is it that nevertheless all reasonable people expect and believe that instances of which they have had no experience will conform to those of which they have had experience? Hume's answer to the psychological version is because of `\emph{custom or habit}'; or in other words, because of \emph{the irrational but irresistible power of the law of association}. We are conditioned by repetition, and without such conditioning mechanism we could hardly survive. %
\end{itemize}
Popper accepts Hume's answer to the logical version of the problem. That is, we certainly are \emph{not} justified in reasoning from one or more instances to the truth of a universal regularity or law (from which we should expect new or unseen instances to abide by). 
Popper adds to it though the important insight that we do are justified in reasoning to the \emph{falsity} of a law like ``All graduate white males are good hires'' 
from a counterinstance (say, some graduate white guy who turns out not to be a good hire), which \emph{if accepted} as such, can disprove or refute it.
In fact, the emerging research community working on `bias of AI' challenges has started to approach issues implicitly related to Hume's logical problem of induction, 
as we will discuss later in \S\ref{sec:related-work} on related work.

Yet, Popper draws attention to a dilemma between Hume's answers. On the one hand, repetition has no power whatsoever as an argument (that is, in reason). On the other hand, it dominates our cognitive life or our `understanding',%
\footnote{And most importantly, for the purpose of this paper, it is at the core of data-driven AI.}
so if we take Hume's answer to the psychological problem `as is,' however intuitive, it leads to the conclusion that our knowledge is of the nature of \emph{rationally indefensible belief} and then can't be distinguished from irrational faith. Popper does \emph{not} accept it at all, and has rather come up with an ingenious alternative. It opens an encouraging landscape for the community working on bias-of-AI challenges, which can't afford staying with Hume's skepticism either --- otherwise AI-based systems may not be rationally defensible. 

Popper's epistemological framework is meant to solve the psychological problem of induction (and also the related formulations, so-called ``the pragmatic problem'') in a way that satisfies ``the principle of the primacy of the logical solution'' (cf. \citeauthor{popper1953}, \citeyear{popper1953}). That is, rather than accepting Hume's irrationalist consequences, we rely on his answer to the logical problem, which is stronger, and transfer it to the psychological one as follows. 
As a first basis, 
Hume's negative result has established for good that all our universal laws or theories remain forever guesses, conjectures, \emph{hypotheses}. Nonetheless, on top of his own (extended) negative result, namely, that `refutation' is a logically valid operation, Popper has developed a \emph{logical theory of preference} --- that is, preference from the point of view of the search for truth. So a purely intellectual preference, for one or another hypothesis, can be drawn from how they stand against \emph{attempts at refutation}. Thereby, one can have purely rational arguments to \emph{prefer} some competing hypotheses to others. It is also possible for our conjectural knowledge to improve. In fact, we can \emph{learn} by conjectures and refutations. Popper's ideas have been seen as offering a full-fledged learning theory \cite{popper1984}, which we leverage on next as for bias mitigation in AI-based systems.


Conveniently put by \citeauthor{popper1984} \shortcite{popper1984} as a learning theory focused on problem solving, Popper's view goes as follows. A problem exists when an observation is contrary to what is expected. This discrepancy stimulates efforts to correct expectancies in order to make them compatible with the previously surprising observation. The newly formulated expectancies remain intact until observations are made that are incompatible with them, at which point they are revised again.
This process of adjusting and readjusting one's expectations so that they agree with the observation is an unending process. However, it is a process that, one hopes, makes expectancies and reality increasingly compatible (Ibid., p.~16). Since it is reasonable (and ethical) to act based upon both the best of our knowledge at present and the best of our critical effort to keep looking for holes or missing points in it, then one can make reasonable decisions by drawing from the best hypothesis available at a given time, namely, the one which is standing better our serious attempts at refutation.%
\footnote{So, the probability of a hypothesis has nothing to do with how much data it has been trained upon by but only with how much it has stood against attempts at refutations (using independent evidence) in comparison with competing hypotheses.}
In sum, we have the  argument that a ``less refuted'' hypothesis should be expected to deliver better results, therefore, preferred. If we are able to collect new, independent evidence over time into an online computational system, then what we have an online learning theory.

There are three key points in this epistemologically-founded theory for online learning:

\begin{itemize}
\item \emph{Acceptance of evidence}. The acceptance of a counterinstance or counterevidence as such (recall our graduate hire counterinstance) is a question of methodological and even ethical nature but not of logical nature. Essentially, scientists have to opt/decide to act according to this \emph{methodological or ethical rule}: if our hypothesis is threatened by one or more counterinstances, we will never save it by any kind of conventionalist stratagem but rather will take seriously the responsibility to review and, if needed, to revise it (cf. \citeauthor{popper1934}, \citeyear{popper1934}; \citeauthor{popper1974}, \citeyear{popper1974}). In AI, this shall not be a problem, as long as a clear-cut separation between training and testing data is preserved; in other words, hypotheses will be tested against a growing body of independent evidence that is accumulated.

\item \emph{Plurality of hypotheses}. Suppose one has only one hypothesis $f$ (say, a binary classifier) to live by, and it has been formed (say, by training) by considering some portion of the world (a dataset $D$). If this hypothesis is threatened by a counterinstance from dataset $D^\prime \!\supset\! D$, then in principle it could be revised in some way (say, retraining) to get better at the problem at hand. 
Yet, if no other hypothesis $f^\prime \neq f$ is available, as new problem instances come in, one has to apply their suspicious hypothesis $f$ anyway. With a portfolio of hypotheses $f \in \mathcal F$, on the other hand, if some hypothesis is continuously refuted for a target problem, the online learning framework may be able to handle it with another, ``more natural'' hypothesis. Therefore, the portfolio as a whole has broader scope to address a variety of problems. In fact, the question of `preference' will arise mainly, and perhaps solely, w.r.t. a set of competing hypotheses \cite[p.~13]{popper1972}. 

\item \emph{Accuracy and fairness}. Popper's epistemological principles are geared for the empirical sciences \cite{Popper:1985}, which shall include the sciences of the artificial such as AI as long as some notion of truth (empirical adequacy) is available. In AI, at a high level of abstraction, we call it \emph{accuracy}. Attempts at refutation should improve knowledge of how `accurate' our hypotheses are, so that preferring one hypothesis over another, after one or more such attempts, is a matter of accuracy. In AI, however, as system outputs are often predictions about human beings, accuracy has to explicitly interact with the value of fairness. For this reason, we will consider that attempts at refutation shall also stress whether the hypotheses embedded in a system are able to make it a fair piece of AI.

\end{itemize}




\section{The Hiring Decision Problem: An Example}
\label{sec:example}

We describe now a  scenario of the hiring decision problem that will be used to illustrate core concepts in our framework. Table~\ref{tab:table1} shows examples of instances as well as their associated outputs according to some decision function~$f$ (a conjectured hypothesis in Popper's sense).

\vspace{-0.45cm}
\begin{table}[h!]
	\begin{center}
		\caption{Simple scenario of the Hiring Decision Problem.}
		\label{tab:table1}
		\begin{tabular}{ c | c | c | c | c   || c} 
			\textbf{ID} & \textbf{Gender} & \textbf{School} & \textbf{City} & \textbf{ZIP} & \textbf{Output} \\
			\hline
			1 & Male & 15  & NYC &  10118  & Yes  \\
			2 & Female & 15 &  Boston & 02110 & Yes\\
			3 & Female & 19 & Chicago & 60603 & No \\
			4 & Male & 10  & Atlanta &  30302  & Yes  \\
		\end{tabular}
	\end{center}
\end{table}
\vspace{-0.25cm}

\noindent\mbox{\textbf{Data model}}.  
Each instance of the problem describes a \textit{job applicant} (or candidate) represented by a vector in~$\mathbb{N} \times G \times \mathbb{N} \times C \times R$. Job applicant~$(i,g,s,c,r)$, to which we may refer as \textit{instance~$i$}, is associated with identification number (ID)~$i$. It is of gender~$g$ in $G = \{$male, female$\}$, has~$s$ years of schooling, has a degree from university~$c$ in $C = \{$NYC, Boston, Chicago, Atlanta$\}$, and resides in an address with ZIP code~$r$ in $R$. We assume that gender is an \textit{explicit sensitive attribute} (i.e., there should be no correlation between the output of a decision function and gender); ID and years of schooling are \textit{non-sensitive attributes}, and ZIP code is an \textit{implicit sensitive attribute} due to correlation with race (a sensitive attribute that does not belong to the data model).

A decision function~$f$ for the problem produces an output~$f(i) \in \{0,1\}$ for each instance~$i$, with $f(i) = 1$ indicating that~$i$ should go to the interview phase (i.e, output is ``Yes'') and $f(i) = 0$ recommending rejection (output is ``No''); that is, $f$ in our case is always a binary classifier.

The proposed framework relies on a ground-truth table qualifying the output produced by the decision functions for each instance~$i$. It consists of the desired output~$f^*(i)$ for~$i$ and of a loss value~$l_i : \{0,1\} \rightarrow [-1,1]$ associated with the correctness of the possible outputs. Namely, $l_i(f) = l_i(f(i))$ is non-negative if~$f(i)$ is wrong and non-positive otherwise. The absolute values of~$l_i$ are correlated with the level of relevance and correctness of the associated decisions. For example, if~$i$ is a  top job applicant, $l_i(0)$ should be close to~$1$ (i.e., this decision should be heavily penalized) and $l_i(1)$ should be close to~$-1$, whereas decisions of~$f$ about borderline candidates may be associated with values~$l_i(f)$ which are closer to zero (e.g., $0.25$ and $-0.25$ for wrong and correct recommendations). Table~\ref{tab:table2} shows feedback values for the instances presented in Table~\ref{tab:table1}. 

\vspace{-0.45cm}
\begin{table}[h!]
		\begin{center}
			\caption{Ground-truth table for sample scenarios.}
			\label{tab:table2}
			\begin{tabular}{ c | c | c | c} 
				\textbf{ID} & \textbf{$f^*(i)$} & $l_i(0) = l_i($No$) $ & $l_i(1) = l_i($Yes$)$  \\
				\hline
				1 &  No  & -1.00  & 1.00  \\
				2 & Yes  & 0.25 & -0.25\\
				3 & Yes  & 0.50 & -0.50\\
				4 & No  & -1.00 & 1.00\\
			\end{tabular}
		\end{center}
\end{table}
\vspace{-0.25cm}

We consider online scenarios, in the sense that new instances are being constantly incorporated to the training set. The arrival of each (batch of) incoming instance(s) corresponds to Popper's attempt at refutation principle.


\noindent\mbox{\textbf{Accuracy and fairness}}. Mistakes made by a decision function~$f$ may be assessed at two levels. At a more fundamental level, one may check whether~$f(i)$ is correct by direct comparison with~$f^*(i)$. This type of error can be identified by inspecting the behavior of~$f$ at a single instance and can be directly interpreted as a \textit{accuracy} problem. This kind of mistake typically takes place  because~$f$ adopted wrong inferences, which may also have a discriminatory aspect; this is the case if $f(i) = 1$, $f(i') = 0$, and instances $i$ and~$i'$ differ solely by gender, for example.

At a higher level, one may identify fairness issues associated with decision function~$f$. Differently from accuracy, fairness relies on the analysis of the behavior of~$f$ for the whole dataset with respect to one or more sensitive attributes. More precisely, we say that~$f$ is unfair with respect to some sensitive attribute~$a$ if there is a significant correlation between its output and~$a$. For example, if among all candidates from NYC  with 20 years of schooling $f$ accepts 80\% of the male candidates and only 60\% of the female candidates, gender discrimination is taking place.

%


\noindent\mbox{\textbf{Baseline and challenges}}. 
The baseline strategy for this problem 
consists of  a single \textit{offline} decision function; a function is offline if its behavior does not change over time as new instances and their feedback values are made available.

A decision function~$f$ may become outdated over time in offline strategies, and even though this issue may be addressed if~$f$ is periodically retrained, 
decision makers typically lack a structured approach to decided when this process is actually necessary (i.e. retraining needs may be aperiodic).  Moreover, there is hardly a universally acclaimed decision function for the hiring decision problem, as different views and guidelines may lead to different conclusions about what the best option may be. In these scenarios, having a portfolio of candidate decision functions 
can be more interesting than having just one.  In short, knowing how to select decision functions and when to update them are key challenges for the baseline strategy, which we try to address in the proposed framework.

\section{A Framework for Bias Mitigation}\label{sec:model}

The proposed framework 
is a two-level system whose  first level consists of a procedure to \mbox{\emph{select}} a decision function  so that a  good balance between exploration and exploitation is achieved,  and whose second level is targeted at the \textit{enhancement} of decision functions based on empirical knowledge about issues with accuracy and/or fairness.
The exposition is illustrated with the hiring decision problem, but the framework is general and may be applied to other scenarios.

\subsection{Decision function selection}


The primary goal of the first level of the framework is to deliver an output for each incoming instance~$i$. For this, it employs and manages a portfolio~$\mathcal{F}$ of candidate decision functions. A key aspect to be addressed  is the identification of a suitable balance between \textit{exploration} and \textit{exploitation}. Exploration refers to testing functions: namely, one wishes to collect as much information about the functions as possible by applying them to several  instances of the target problem and having feedback on their accuracy. Conversely, exploitation refers to utility maximization: given that some function~$f \in \mathcal{F}$ typically delivers better results than some other function~$f' \in \mathcal{F}$, rational decision makers are more likely to choose~$f$ instead of~$f'$.  As the ground-truth table is unknown by the time decisions are made, a reasonable goal for the selection component is to achieve  relative performance guarantees, such as a bound on the ratio between the number of wrong decisions made by the framework and the number of mistakes made by the best function in~$\mathcal{F}$.\footnote{This metric is typically referred to as \textit{regret} in statistical decision making theory.}
  
Similar problems can be identified in several fields, such as game theory, geometry, operations research, and statistical decision making. Recently, the work of~\citeauthor{arora2012multiplicative} provided   an attempt 
to unify the proposed approaches within the same algorithmic framework, referred to as the Multiplicative Weights Update method (MWU) (\citeyear{arora2012multiplicative}). In general lines, MWU works as follows. Given a set~$\mathcal{F}$ of decision functions  for a target problem~$\mathcal{P}$, each candidate function~$f \in \mathcal{F}$ is  assigned a weight~$w_f > 0$. The value of~$w_f$ is initially set to 1 for each~$f$ and is modified after the arrival of each new instance~$i$ according to the correctness of~$f(i)$, as we describe next. Anytime, terms  $p_f \gets w_f/\sum_{f' \in \mathcal{F}}w_{f'}$ give a probability distribution~$D$. 
As each instance~$i$ is given as input to the framework, 
a \textit{selection phase} takes place, in which a decision function~$f^c \in \mathcal{F}$ is selected by sampling according to~$D$ (i.e., $f$ is drawn with probability~$p_f$). All functions $f \in \mathcal{F}$ are applied to instance~$i$ so that the result $f(i)$ is known for all, but only 
$f^c(i)$ is returned by the framework. 

Once entries~$f^*(i)$ and $l_i$ of the ground truth are generated, the weight~$w_f$ of each decision function~$f$ is updated to reflect changes on the empirical level of confidence the system has on~$f$. A possible expression is  $w_f \gets w_f\, [1-\eta\, \ell_i(f(i)) \,]$, where $0 < \eta \leq \frac{1}{2}$ is a parameter of the problem. In addition to actual instances, synthetic cases may be generated to check whether the model has  discriminatory behavior with respect to some explicit sensitive attribute; e.g., if~$i$ is a real instance and~$i'$  is a synthetic instance differing from~$i$ only in the gender attribute, 
discrimination takes place if $f(i) \neq f(i')$. 
The same procedure may be employed with non-sensitive attributes, 
 thus supporting the identification of absurd behavior and/or indirect discrimination.\footnote{For example, ZIP code may be correlated with race.}
As~$w_f$ becomes small,  the probability with which decision function~$f$ is selected  decreases. Weight~$w_f$ becomes small if~$f$ makes wrong decisions for several instances (associated with high penalties), so~$w_f$ can also be interpreted as a proxy for  the quality of~$f$ as a decision function for problem~$\mathcal{P}$. Moreover, 
procedures for the identification of unfairness 
may also be periodically employed by the framework. Whenever~$w_f$ becomes too small (e.g., goes below some threshold value~$\tau$) or unfair behavior is identified, $f$ is removed from~$\mathcal{F}$ and given as input to
the Decision Function Enhancement procedure on~$f$, which we describe next.

\subsection{Decision function enhancement}



The second level of the framework is targeted at the enhancement of a decision function~$f$ that has been removed from the portfolio~$\mathcal{F}$ in the first level. In the context of AI systems, such adjustments will ideally address the mistakes that made~$f$ unacceptable for the target problem. In scenarios where decision makers have no control over~$f$, the only alternative is to report the issues to the provider of~$f$; otherwise, the following approaches can be considered. 


\noindent\mbox{\textbf{Retrainable Black-Box decision functions}}  In scenarios where~$f$ 
is a black-box model that can be retrained, a possible  strategy consists of the incorporation to~$f$'s training set of instances  for which~$f$ delivered wrong decisions. The idea of this approach is to employ knowledge about previous mistakes made by the algorithm, with the hope that retraining will be sufficient to mitigate or even eliminate some of the issues that have been observed. However, there is no guarantee that~$f$ will eventually become completely free of all mistakes, even after a large number of iterations. 
Moreover, it is not clear how the incorporation of new instances to the training set may unfairness.


\noindent\mbox{\textbf{Non-black-box decision functions}} 
If~$f$ can be modified, more effective strategies may be employed. In particular, constrained models become specially interesting in these scenarios, as restrictions may help in the reduction of undesired effects such as unfairness. For example, in our scenario, if~$f$ is a margin-based classifier, one may employ a constrained optimization model 
to bound the correlation between subsets of attributes and the output of~$f$ for instances of the training set~(see e.g., \citeyear{zafar2017fairness}).

If all relevant constraints are known a priori and their inclusion in the underlying model leads to a computationally tractable formulation, the desired decision function can be directly constructed. This is typically not the case, though, and this is where \textit{cut-generation} algorithms, which have been strongly investigated in the literature of constrained optimization and constrained logic programming, can play a key role. In the mathematical programming literature, cut-generation algorithms are useful in scenarios where one does not work with a full representation of the decision function~$f$, but with a \textit{relaxation}~$f^r$ of~$f$, which lacks some of the original constraints and conditions. Relaxations are used by convenience (e.g., in scenarios where the number of constraints is too large and having a smaller formulation is computationally more interesting) and by necessity (e.g., when the relevant constraints are unknown a priori), and both cases may apply in scenarios where discrimination may take place.  For instance, an attribute that was initially considered non-sensitive may be a proxy for some other sensitive attribute that is not explicitly represented in the data model (e.g., in our case, ZIP codes and race), and this correlation may  be identified only after the observation of new instances. In this case, a function~$f$ that ignored this aspect previously  can be fixed by the inclusion of a new constraint to the formulation, capable of mitigating this type of bias.

\citeauthor{zafar2017fairness} employ a single subset containing all sensitive attributes, but a more comprehensive approach would require the inspection of all possible combinations of sensitive attributes. For our example, given that there are two sensitive attributes (Gender and ZIP), three subsets could be inspected. More generally, in scenarios with~$s$ sensitive attributes, $O(2^s)$ subsets may be created, which may be prohibitively large in scenarios with relatively high number of dimensions. The use of relaxations combined with cut-generation are convenient in these cases, as the resulting models would be otherwise computationally intractable.

\section{Related Work}\label{sec:related-work}

Technical frameworks supporting the design of systems with socially acceptable behavior have been proposed in the literature. For instance, \citeauthor{rossi2016preferences} and \citeauthor{greene2016embedding} discuss the challenges involved in the creation of ``ethic algorithms'' and how they can be addressed by techniques in preference modeling and reasoning, multi-agent systems, constraint programming, constraint-based optimization, etc. (\citeyear{rossi2016preferences,greene2016embedding}). These contributions assume that systems of preferences can be captured and formalized, so the main challenge lies in aggregating and consolidating different preference lists in scenarios where two or more lists may be considered, and eventual conflicts will then be resolved in order to have an ethical outcome. 
The technical framework we have proposed has a different approach and follows different design guidelines. First, our main goal is to devise a system capable of dealing with 
hypotheses and problems for which 
discrimination and unfairness may be observed 
and could be mitigated. These challenges emerge at a more fundamental level than those addressed by Rossi et al., as  
ethical issues related with bias, unfairness, etc. do not appear  
due to differences in preference, 
but due to issues with the data collection process, lack of algorithmic transparency, etc..

Bias on data has been discussed by in many contexts from social data used to support decision or to characterize human phenomena 
\cite{surveyDataBias} to solutions focus on discrimination discovery \cite{Ruggieri:2010:DMD:1754428.1754432} to discrimination prevention \cite{zafar2017fairness}.

In addition to the work of \citeauthor{zafar2017fairness}, others have also proposed constrained models to enforce fairness. For instance, \citeauthor{celis2017ranking} investigate the problem of ranking a set of items subject to families of fairness constraints~(\citeyear{celis2017ranking}). Group-fairness constraints have also been considered for multi-winner voting problems~\cite{celis2017voting}. For an overview of the Multiplicative Weights Update method and its relationship with several similar techniques in other areas, see e.g.~\cite{arora2012multiplicative}. Recently, \citeauthor{celis2017personalization} employed a similar algorithmic  framework (i.e., based on stochastic contextual bandits) to control bias and discrimination in systems of online personalized recommendation (\citeyear{celis2017personalization}). The literature in cut-generation algorithms is vast, with celebrated results in the area of mathematical programming for problems with very special structural properties (see, e.g., \cite{grotschel2012geometric} and \cite{hooker2003logic}.



\section{Discussion and Future Work}
\label{sec:conclusions}

In this paper, we discuss AI challenges enlighted by Karl Popper's epistemological principles and modeled into well-known concepts from (theoretical) computer science and optimization. We have pointed out connections between bias of AI and the problem of induction, focusing on Popper's contributions after David Hume's. We have seen that it offers a logical theory of preferences, that is, one hypothesis can be preferred over another on purely rational grounds after one or more attempts at refutation. Based on these ideas and their suit into a learning theory, we have proposed a two-level computational framework. Decision functions (hypotheses in Popper's sense) are selected while keeping a good balance between exploration and exploitation. Those presenting low accuracy and unfair behavior are continuously enhanced.  



In future work, we plan to carry out a large study of the hiring problem, followed by an extensive evaluation of our proposed framework. Our goal is to add empirical results to our epistemological argument, which will also be deepened and improved as we further develop our research on bias and philosophy of AI.  
As a final remark we recall that AI systems, as they rely on human-generated data, will always be susceptible to bias, sometimes prejudice. 
Solutions such as ours, modeling epistemological (and ethical) principles into a computational framework for mitigating bias, shall be complemented in order to take into account social, legal and other bias-of-AI challenges. 

\newpage
\bibliographystyle{aaai}
\bibliography{references}

\end{document}